\documentclass[letterpaper, 10 pt, conference]{ieeeconf}  

\pdfoutput=1                                              
\IEEEoverridecommandlockouts                              

\usepackage{cite}
\usepackage{amsmath,amssymb,amsfonts}
\usepackage{algorithmic}
\usepackage{graphicx}
\usepackage{caption}
\usepackage{subcaption}
\usepackage{textcomp}
\usepackage{xcolor}
\usepackage{siunitx}
\usepackage{float}
\usepackage{svg}
\usepackage{amsmath,amssymb}
\usepackage{hyperref}
\usepackage{verbatim}
\usepackage[normalem]{ulem}

\graphicspath{ {images/} }

\def\BibTeX{{\rm B\kern-.05em{\sc i\kern-.025em b}\kern-.08em
    T\kern-.1667em\lower.7ex\hbox{E}\kern-.125emX}}

\title{\LARGE \bf SCOUT: Socially-COnsistent and UndersTandable Graph Attention Network for  Trajectory Prediction of Vehicles and VRUs}

\author{S. Carrasco$^{1}$, D. Fern\'{a}ndez Llorca$^{1,2}$ and M. A. Sotelo$^{1}$
\thanks{$^{1}$ Computer Engineering Department, Polytechnic School, University of Alcal\'a, Madrid,  Spain. \{sandra.carrascol,  david.fernandezl, miguel.sotelo\}@uah.es \newline
$^{2}$ European Commission, Joint Research Center, Seville, Spain. david.fernandez-llorca@ec.europa.eu}%
}

\begin{document}

\maketitle
\thispagestyle{empty}
\pagestyle{empty}

\begin{abstract}
Autonomous vehicles navigate in dynamically changing environments under a wide variety of conditions, being continuously influenced by surrounding objects. Modelling interactions among agents is essential for accurately forecasting other agents’ behaviour and achieving safe and comfortable motion planning. In this work, we propose SCOUT, a novel Attention-based Graph Neural Network that uses a flexible and generic representation of the scene as a graph for modelling interactions, and  predicts socially-consistent trajectories of vehicles and Vulnerable Road Users (VRUs) under mixed traffic conditions. We explore three different attention mechanisms and test our scheme with both bird-eye-view and on-vehicle urban data, achieving superior performance than existing state-of-the-art approaches on InD and ApolloScape Trajectory benchmarks. Additionally, we evaluate our model's flexibility and transferability by testing it under completely new scenarios on RounD dataset. The importance and influence of each interaction in the final prediction is explored by means of Integrated Gradients technique and the visualization of the attention learned.
\end{abstract}

\section{Introduction}

Predicting traffic participants' trajectories is of major importance  in autonomous driving applications, since it allows the controller to plan ahead the motion of the vehicle, avoiding collisions and making better driving decisions. In this work, we aim at socially-aware and socially-consistent vehicles and VRUs trajectory forecasting and interaction understanding.  
Accurately forecasting the motion of surrounding agents  is an extremely complex and challenging task, considering that many factors can affect the future trajectory of an object. First of all, the variety and complexity of road scenes is immense and traffic scene dynamics can be extremely different among different, or even similar, scenarios. Therefore, one major challenge of developing prediction methods is to find comprehensive and generic representations for all common scenarios that can be encountered in the real world. Moreover, although deep learning based models have shown incredible forecasting abilities, it would be desirable to retain structural information and explainability, instead of relying on the blackbox nature of these models. 

\begin{figure}[t]
    \centering
    \includegraphics[width=\linewidth]{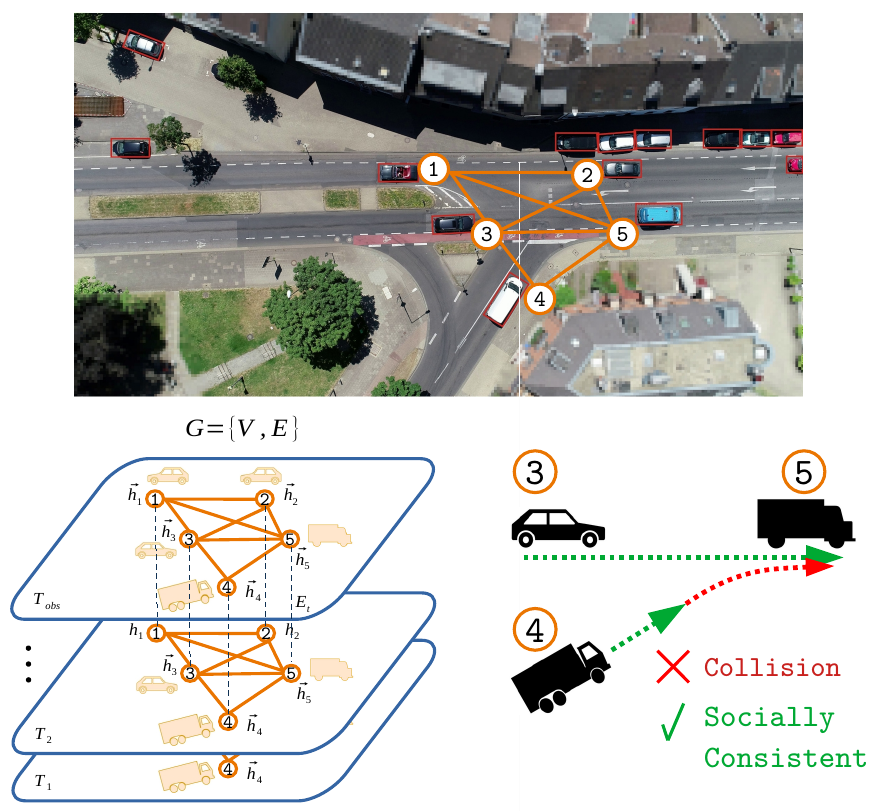}
    \caption{Spatio-temporal graph of an interaction scenario with 5 agents involved. Our approach predicts trajectories using an attention mechanism in a socially-consistent way.}
    \label{fig:Intro}
    \vspace{-2mm}
\end{figure}

Motion forecasting must be socially-aware, i.e. it must consider the past trajectories and intentions of surrounding agents. However, one major open question in the field of  motion forecasting is how to model such interactions among traffic agents. Understanding how the ego-vehicle actions might influence other actors’ behaviors  is essential for safe and comfortable  motion planning of self-driving vehicles.  Additionally,  these predictions must be consistent among vehicles and non-overlapping.  This can only be achieved by deeply understanding the scene dynamics and the essence of interactions among traffic participants. 

On the other side, most deep learning based models used for trajectory forecasting operate on data of a fixed size and a fixed spatial organization, which impedes to obtain a general representation for inputs and outputs such that they can be flexible to the number and type of agents as well as transferable under different scenarios. In our work, we propose to tackle these problems harnessing the power of graph neural networks by modelling each  traffic agent as a node and possible interactions between them as edges, obtaining a high-level representation of the traffic scene as a graph (see Fig. \ref{fig:Intro}). In the light of the above mentioned, our main contributions can be summarized as follows:

\begin{itemize}
    \item \textbf{Socially-aware}: we propose \textit{SCOUT}, a generic graph-based formulation for modelling traffic interactions, where the influence of interactions among vehicles is modelled as an additional element that is dynamically learnt during the training phase in a semi-supervised way, following an attention mechanism.

\item \textbf{Socially-consistent}: trajectory forecasts are learnt by incorporating the overlap of future trajectories as an element to be minimized during training.

\item \textbf{Flexible and transferable}:  our model works for a variety of number and type of road agents, while proving transferability among different scenarios. 

\item \textbf{Urban dataset}: this work has been evaluated with three real-world urban datasets, in which numerous interactions between various road agents occur simultaneously. 

\item \textbf{Interaction understanding}: the exploration of attention learned by our model sheds light on the interpretation of the influence of vehicle interactions on the final prediction.

\end{itemize}


\section{Related work.}
\label{sec:relatedwork}

Trajectory prediction has been extensively studied over the last decades, since it is a key feature for autonomous driving. This has led to a plethora of competitive models and algorithms. 
Recently, deep learning based methods have emerged for maneuver classification \cite{Izquierdo2019} and trajectory prediction \cite{izquierdo2020vehicle}. Specially, Recurrent Neural Networks (RNNs) such as LSTMs have been widely used in the field.  However, most approaches do not take inter-object interactions into account and are therefore limited.

More recent works have proposed new schemes to explore this limitation. In earlier works such as SocialLSTM \cite{dasgupta_cvpr16} and Convolutional Social Pooling (CSP) \cite{DBLP:journals/corr/abs-1805-06771}, interaction among smart agents are implicitly modeled by the “social pooling” operation. 
In \cite{DBLP:journals/corr/abs-1812-04767}, the authors presented a LSTM-CNN hybrid network which take into account heterogeneous interactions modelling each road agent with an LSTM. They create an horizon map by pooling together the agents' hidden states and define a neighborhood map. A major drawback of this scheme is that it requires high computational power.
Most works such as \cite{DBLP:journals/corr/abs-1907-07792} or \cite{8848853} work with a fixed number of closest road entities for studying the influence of interactions, which results in a lack of flexibility. 

Graph Neural Networks has recently emerged as a solution to handle different input sizes and achieve invariance to input ordering, since it processes strong inductive biases \cite{battaglia2018relational}, \cite{diehl2019graph}. \cite{DBLP:journals/corr/abs-1907-07792} utilizes a graph to represent the interaction among nearby agents and uses an encoder-decoder LSTM to make predictions. However they treat each surrounding agent equally by using a binary adjacency matrix to construct the graph. As opposed to modeling interaction implicitly, \cite{Fetaya2018NeuralRI} models the interactions in dynamical systems as latent edge types of an interaction graph, which are learned in an unsupervised manner.  \cite{Casas2020SpAGNNSG} models the multi-agent interactions with a Graph Neural Network, where the node features are derived from the combination of cropped map information and the upstream features extracted from the sensor data. 

Notwithstanding, most of these models are too complex and lack of explainability due to their black-box nature. We propose a simplistic model by thinking about traffic agents and their interactions as nodes and edges in a graph, using an attention mechanism to extract relevant features encoding these interactions and a simple feed-forward network as a behavior prediction model for each node.

\section{Method}
 \label{sec:method}
\begin{figure*}[th]
    \centering
    \includegraphics[width=\linewidth]{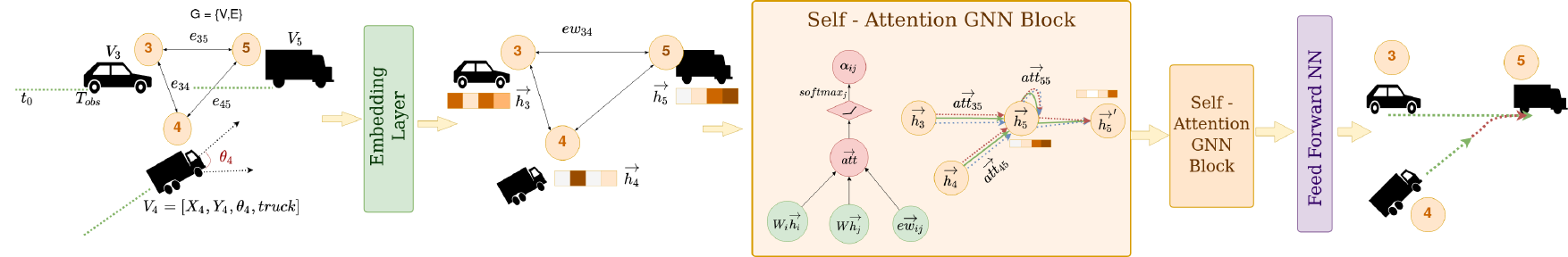}
    \caption{The SCOUT model. Given T timesteps of observed trajectories, we construct the spatio-temporal graph  G=(V,E), where V contains x,y-coordinates, heading and object type of each agent in the scene. Three attention heads aggregates node's features that are subsequently forwarded to a second graph attention layer. Finally, the extracted features are fed to a Feed-Forward layer for the final prediction.}
    \label{fig:Model}
    \vspace{-3mm}
\end{figure*}

To solve the limitations faced by existing approaches, we propose \textit{SCOUT}, a novel deep learning graph-based model for motion forecasting (see Fig. \ref{fig:Model}). In this section we describe our approach and implementation details.

\subsection{Problem Formulation}
\label{Backbone}

The goal is to estimate the future position of all road agents from their past trajectories. Input features $X$ define observed trajectories over $t_{h}$ time steps $t_h \leq T_{obs}$ of all traffic agents in a scene, where $X^t=[X^{t}_{1}, …,X^{t}_{N}]$ for N agents in the scene. Our goal is to forecast future trajectories $\hat{Y}_t = [\hat{Y}^t_1,...,\hat{Y}^t_N]$ for all observed participants over the predicted horizon $T_{obs} < t \leq T_{pred}$. 
The input features of object $i$ at time t are defined as $X^t_i = (x^t_i, y^t_i, \sigma^t_i)$, representing the cartesian coordinates and the heading of the object at that timestep. During training, the predicted positions are compared to ground truth future trajectories $Y$, where $Y_i^t = (x_i^t, y_i^t)$.

\subsection{Graph Construction}

\subsubsection{Feature Representation} 

The first task to be solved is to construct the features in such a way that they can be fed to a graph in an efficient way. Having N agents in the last observed frame ($t_{obs}$), we build a 3D matrix of shape $(N, T_h, C)$, where $T_h$ is the number of frames in the history horizon and $C$ the number of input features.

\subsubsection{Graph Representation}
\label{sec:graph_construction}

We propose to model interactions among traffic agents through the use of a graph representation, as researchers have done for social networks. Being the undirected graph defined as G = \{V,E\}, each node $V_i$ in the node set $V = \{V_1,...,V_N\}$ corresponds to a traffic participant. $V_i$ is defined as $V_i = \{v_i^t | t=1,..,t_h\}$, being $v_i^t$ the feature vector of node $i$ at time $t$.
Now the question at hand is how to construct such  graph, i.e. how to define $E$. We obtain an adjacency matrix from the last observed frame $t_{obs}$. All agents involved in this frame are represented by a node in the graph and are connected by two types of edges. First, we connect each agent $v_i$ with itself through a \textit{temporal edge} $e_i^t$. Hence, $E_t = \{ v_i^t, v_i^{(t+1)}\}$ represents the trajectory history of each node over $t_h$. Secondly, each node is connected with all its surrounding agents that fall inside a radius of 20 meters, the node set $D$. Therefore, the \textit{spatial edge} set is defined as $E_s = \{e_{ij} | (v_i,v_j \in D)\}$. The binary adjacency matrix will be a symmetric matrix of size $(N,N)$:

\begin{equation} \label{eq:1}
   A = \left \{ \begin{matrix} 1 & \mbox{if } \mbox{ edge $(v_i^t,v_j^t) \in D$ }
\\ 0 & \mbox{otherwise }\end{matrix}\right.  
\end{equation}

In addition, we evaluate the use of a weighted adjacency matrix. Considering that a traffic participant is more influenced by close neighbours, we compute the edge weights as an initial fixed attention score, where closer neighbors are assigned a higher value. This value is computed by a kernel function for each edge, which is defined as the inverse of the $L_2$ norm of the relative distances (Eq. \ref{eq:1.2}) , with self-loops set to 1. 

\begin{equation} \label{eq:1.2}
   A = \left \{ \begin{matrix} \frac{1}{||d^t_i - d^t_j||_2} & \mbox{if } \mbox{ edge $(v_i^t,v_j^t) \in D$ }
\\ 0 & \mbox{otherwise }\end{matrix}\right.  
\end{equation}

\subsection{Proposed Model}
\label{sec:model}

Graph Convolutional Networks (GCN) \cite{kipf2017semisupervised} are an approach for semi-supervised learning on graph structured data based on the principles of Convolutional Neural Networks to operate directly on graphs in an efficient manner. Analogous to CNNs, GCNs perform Template Matching, applying the same operation on all nodes. Mathematically, for each node we have:

\begin{equation} \label{eq:2}
 h_i^{l+1} = \sigma( \displaystyle\sum_{j \in \mathcal{N}(i)} \frac{1}{c_{ij}} W^{l}h_j^{l}  )
\end{equation}

where $\mathcal{N}(i)$ is the set of its one-hop neighbors and $c_{ij} = \sqrt{\mathcal{N}(i)}\sqrt{\mathcal{N}(j)}$ is a normalization constant based on graph structure. $\sigma$ is an activation function and  $h_j^{l}$  and $W^{l}$ are node $j$'s feature vector and the template vector at layer $l$, respectively. 
The vectorial representation is:

\begin{equation} \label{eq:3}
 H^{l+1} = \sigma( \tilde{D}^{-\frac{1}{2}} \tilde{A} \tilde{D}^{-\frac{1}{2}}  H^l  W^l  )
\end{equation}

 $\tilde{D}$ and $\tilde{A}$ are the degree and adjacency matrices of the graph, respectively, with a $(n,n)$ size. 
 $ H^{l+1}$ is the activation function at layer $l$ with dimensions $(n,d)$. This is equivalent  to a first-order approximation of a localized spectral filter, where 
 $l$ layers consider exactly $l$ hops in the neighborhood. Consequently, computational complexity scales linearly in the number of edges and can take topologically distant vehicles into account. 
This formulation is also independent of the graph size, since all operations are done locally.
One drawback of this scheme is its isotropic nature. 
Anisotropy can be introduced by giving a different weight to each neighbor.  We tried three aggregation approaches to introduce this anisotropy:

\begin{enumerate}
    \item \textbf{Fixed attention-weight.}
    
     We compute the edge weights for each pair of nodes $e_{ij}$ using as kernel function the inverse of the $L_2$ norm of their relative distance, setting self-loops to 1. This approach takes the assumption that traffic agents are more influenced by nearby neighbours and assigns a fixed attention score for each interaction between nodes.
    In this way, the message function for each node would be:
    
      \begin{equation} \label{eq:4}
         m_i^{l+1} = \sigma( \displaystyle\sum_{j \in \mathbb{N}(i)} \frac{e_{ij}h_j^{l}}{c_{ij}+1} ),
      \end{equation}

    However, this formulation presents one big disadvantage, as it treats the ego node’s own features equally as the ones of its neighbors, applying the same weights. This could significantly decrease the performance of our prediction system.  Hence, we introduce a residual weight matrix defining a rotation on the ego node’s features. 

    \item \textbf{Attention mechanism in the neighbourhood aggregation function.}
    
    Graph Attention Network \cite{article} uses weighting neighbor features with feature dependent and structure-free normalization, in the style of attention. 
    We introduce the attention mechanism as a substitute for the statistically normalized convolution operation. In this case, given a source (neighbourhood features) and a \textit{Query} (central node), the output value would be the weighted sum of all the \textit{Values}, being the weight values the correlation between the \textit{Query} and the \textit{Key}.  
    Mathematically, the attention score for each pair of nodes is calculated as follows:
    
    \begin{equation} \label{eq:7}
    \alpha_{ij}^{l} = Softmax ( LeakyReLU ( a^{l^T} · z_{ij} ) )
     \end{equation}
     \begin{equation} \label{eq:7.2}
      z_{ij}^{l} = [W^l h_i^l ||W^l h_j^l || e_{ij} ] 
     \end{equation}
      
    Eq. \ref{eq:7} computes a pair-wise un-normalized attention score between two nodes. First, we applied a linear transformation of $h_i^l$ and $W^l$. It then concatenates the features of the two neighbours and their corresponding edge and computes the dot product with a learnable weight vector ($a^{l^T}$), applying finally a LeakyReLU activation. Then, softmax is applied to normalize the attention scores on each node’s incoming edges. The aggregation of the neighbours’ embeddings is performed similarly to GCN, scaled by the attention weights ($\alpha_{ij}$), as shown in Eq. \ref{eq:8}.
    
    \begin{equation} \label{eq:8}
         h_i^{l+1} = \sigma(\displaystyle\sum_{j \in \mathbb{N}(i)} \alpha_{ij}^{l} W^{l} h_j^{l} )
    \end{equation}
    Analogous to multiple channels in CNNs, GAT introduce multi-head attention to enhance the model capacity and to stabilize the learning process, i.e. using k  differently-parameterized attention heads and concatenating the result, which allows each head to focus on a subset of features or nodes.
    
\item \textbf{Edge gating mechanism.}

This approach can be seen as a softer attention process than the sparse attention mechanism used in the previous point. Here, edges have a feature representation that evolves in each layer and is learned dynamically during training. Edge features are computed as defined in Eq. \ref{eq:1.2}. The hidden representation of each node is computed as follows:
 \begin{equation} \label{eq:9}
         h_i^{l+1} = h_i^l + \sigma(Ah_i^l + \displaystyle\sum_{j \in \mathbb{N}(i)} \eta(e_j) \odot Bh_j)
    \end{equation}
$ \eta(e_j) \odot Bh_j $ denotes the sum of elementwise multiplication of rotated inputs features and a gate. Here, the gate term is crucial, since it allows to modulate the incoming representations based on the edge features. The gate term is computed by Eq. \ref{eq:10}. It is a normalized sigmoid computed with the incoming edge representation, defined in Eq. \ref{eq:11}, where C, D and E are learnable matrices.
\begin{equation} \label{eq:10}
         \eta(e_j) = \frac{\sigma(e_j)}{\displaystyle\sum_{k \in \mathbb{N}(i)} \sigma(e_k)}
    \end{equation}
    \begin{equation} \label{eq:11}
    e_j^{l+1}=  e_j^l + ReLU(Ce_j^l + Dh_j^l + Eh_i^l)   
    \end{equation}

\end{enumerate}

The proposed model is composed by one first embedding layer for transforming the input node and edge features, two graph layers, defining exactly two hops in the neighbourhood, and finally a feed-forward layer that acts on each node independently, which allows for a better decoupling of the feature extraction and the final forecasting task.

\subsection{Implementation Details}

Our model was implemented using PyTorch Lightning \cite{falcon2019pytorch}, a lightweight PyTorch wrapper for high-performance AI-research and  Deep Graph Library \cite{dgl}, a Python package for deep learning on graphs. We trained our model using AdamW optimizer with a learning rate of 1e-4 and using a batch size of 256 samples. Model's weights were initialized using Kaiming initialization. A weight decay of 0.01 and a dropout probability of 0.25  were used to control overfitting. Additionally, we applied dropout with a probability of 0.6 to the attention function, to make sure the attention is properly learned while maintaining its genaralizability. The  training  was  performed  as  a  regression  task,  being  the overall loss computed as:

\begin{equation} \label{eq:13}
     Loss = \frac{1}{T_{pred}} \displaystyle\sum_{t=1}^{T_{pred}} L_\delta^t (1 + \alpha*p\_overlap) + \beta L_\delta^{T_{pred}} 
\end{equation}

\begin{align}\label{eq:13.2}
loss_\delta(y,\hat{y}) = \left\{ \begin{array}{cl}
\frac{1}{2} \left[y_i - \hat{y}_i\right]^2 & \text{for }|y_i - \hat{y}_i| \le \delta, \\
\delta (\left|y_i - \hat{y}_i|-\frac{\delta}{2}\right) & \text{otherwise.}
\end{array}\right.
\end{align}

\begin{equation} \label{eq:14}
     L_\delta^t = \frac{1}{n}\displaystyle\sum_{i=1}^{n} loss_\delta  
\end{equation}
where \textit{$L_\delta$} (\ref{eq:13.2}) is a combination of Mean Square Error (MSE) and Mean Absolute Error (MAE), also known as Huber Loss. It combines good properties from both MSE and MAE. On the one hand, it overcomes the constant large gradient that derives from MAE even for small loss values, which is detrimental for learning. On the other hand, MSE behaves nicely in this case and will converge more precisely at the end of training, however it is much more sensitive to outliers. Huber loss addresses this issue by using MAE when error is high and becoming quadratic when error is small. Hyperparameter $\delta$ represents the boundary between both functions.
$T_{pred}$ is the predicted horizon, $L_\delta^t $ is the Huber loss at time t between output $Y$ and ground truth $\hat{Y}$ positions and $n$ is the number of nodes or road agents present in the actual frame. The factor $(1 + \alpha*p\_overlap)$ penalizes the overlap of predicted trajectories during training, thus learning in a socially consistent way. We compute the overlap percentage (\textit{p\_overlap}) between the estimated output for all the nodes in the neighbourhood  and add it as a penalty term to the loss function.
Finally, the rightside of Eq. \ref{eq:13}, $\beta L_\delta^{T_{pred}}$, allows us to minimize Final Displacement Error (FDE). Please, refer to the ablation study in section \ref{sec:ablation} for more details. 

\section{Experimental Evaluation.}
\label{sec:results}

In this section we discuss and analyse the results obtained from our experimental evaluation. 

\subsection{Datasets}

In order to validate the capability of the proposed scheme, we conduct several experiments comparing our results with a number of baselines using three real-world datasets with mixed-traffic data. On the one side, InD \cite{inDdataset} and RounD \cite{rounDdataset} datasets contain public traffic data (position, velocity, acceleration and type of object) recorded from a bird-eye-view perspective at the two most relevant and challenging type of traffic scenarios: urban unsignalized intersections and roundabouts. On the other side, we evaluate our system with  an on-road collected dataset, the ApolloScape Trajectory Dataset \cite{ma2019trafficpredict}. This allows us to evaluate our model against datasets of a completely different nature.

\subsubsection{InD Dataset \cite{inDdataset}} The inD Dataset consists of 10h of recorded trajectories at 4 different German intersections in a top-down view. 
All four intersections are unsignalized and contain walkways. Apart from that, they differ in terms of shape, number and types of lanes, right-of-way rules, traffic composition and kind of interaction. 5 types of agents are involved, which allows us to test our architecture for having heterogeneous traffic agents which show completely different behaviors and interactions. We subsample the data, captured at 25Hz, to 2.5Hz, obtaining sequences of 8 steps for observations and  12 steps for prediction.
    
\subsubsection{RounD Dataset \cite{rounDdataset}} it contains 6h of naturalistic road user trajectories recorded at German roundabouts, presenting a wide variety of complex and dense interactions.   This dataset is used to examine the transferability of our system, testing our model learned under the roundabout scenario against the inD test set. Please, refer to section \ref{sec:transfer} for more details.  

\subsubsection{ApolloScape Trajectory Dataset \cite{ma2019trafficpredict}}  This dataset is collected on-road under various   traffic densities  during rush hours in Beijing. It contains highly complicated heterogeneous traffic flows, with a total of 53 minutes of training sequences and 50 minutes of testing sequences captured at 2 fps.    This dataset is particularly challenging, given the limited amount of data and the scenarios covered, where various types of traffic agents with different behaviors and speeds create additional challenges. We train the model using training sequences and submit the results on the testing sequences to the ApolloScape website for evaluation. In this challenge, they evaluate the predicted position in the next 3s (6 frames) given 3s of observation. 

\subsection{Metrics: }

The metrics used for evaluation are \textit{Average Displacement Error (ADE)} and \textit{Final Displacement Error (FDE)}.
ApolloScape Trajectory challenge uses the weighted sum of ADE (WSADE) and FDE (WSFDE) as metrics to differenciate among different agents types by giving a different weight to each type related to reciprocals of their average velocities, i.e. $D_v = 0.20, D_p = 0.58$ and $D_b = 0.22$ for vehicles, pedestrians and bicycles respectively.

\subsection{Ablation Study}
\label{sec:ablation}

We perform an ablation study to understand the contribution of each component to the overall system. Table \ref{table:ablation} shows the relative results w.r.t. to best configuration, taking into account the  parameters listed below. Please, note that \textit{T} stands for True and\textit{ F }stands for False.

\subsubsection{\textbf{Aggregation Function}}

We test: (A)  Fixed attention weight, (B) Attention mechanism and (C) Edge gating mechanism. 
Best results are shown for each model type after adjusting all the involved parameters. Configuration (B) achieves the best results probably due to the strong attention mechanism involved and the multihead, which reinforce it even further (absolute values are shown in table \ref{table:agents}). We perform the remaining study starting from this configuration, analysing how the different features affect the final error.

\subsubsection{\textbf{Residual connection} (RC)}

In order to calculate the hidden representation for the nodes, we add the input representation fed to each of the graph layers. The model experiments noticeable decrease in performance when this residual connection is removed.

\subsubsection{\textbf{Residual weight} (RW)}

Table \ref{table:ablation} shows that removing the residual weight  increases  the  prediction  error  by  at  least  11\%.  This  supports our belief that there  is  a  clear  difference  between  neighbours and  the  ego  node  in  this  task and should be treated accordingly.

\begin{table}[tp]
    \caption{Ablation study. Results show the \% of error increment w.r.t. best configuration.}
    \begin{center}
        \begin{tabular}{ccccccc|c}
        \multicolumn{1}{p{0.5cm}}{\centering \textbf{Agg.} \\ \textbf{Fcn.}}
        & \multicolumn{1}{p{0.5cm}}{\centering\textbf{RC}} 
        & \multicolumn{1}{p{0.5cm}}{\centering\textbf{RW}}  
        & \multicolumn{1}{p{0.5cm}}{\centering \textbf{Att.} \\ \textbf{Heads}}  
        & \multicolumn{1}{p{0.5cm}}{\centering\textbf{FC}}
        & \multicolumn{1}{p{0.5cm}}{\centering\textbf{$\beta$}} 
        & \multicolumn{1}{p{0.5cm}}{\centering\textbf{SC ($\alpha$)}}   
        & \multicolumn{1}{p{1cm}}{\centering\textbf{ADE/FDE} } \\ \hline\hline
        A & T  & T  & 3 & T & 1 & 5 & +37\% \\ \hline
        B &  T & T  & 3 & T  & 1 & 5 & Best \\ \hline
        C &  T & T  & 3 & T  & 1 & 5 &  +18\% \\ \hline
        B & F  & T  & 3 & T & 1 & 5 &  +30\% \\ \hline
        B & T  & F  & 3 & T & 1 & 5 &  +11\% \\ \hline
        B & T  & T  & 1 & T & 1 & 5 &  +25\% \\ \hline
        B & T  & T  & 4 & T & 1 & 5 &  +16\% \\ \hline
        B & T  & T  & 3 & F & 1 & 5 &  +14\% \\ \hline
        B & T  & T  & 3 & T & 0 & 5 &  +2 / +7\% \\ \hline
        B & T  & T  & 3 & T & 2 & 5 &  +5 / -2\% \\ \hline
        B & T  & T  & 3 & T & 1 & 0 &  +17\% \\ \hline
        B & T  & T  & 3 & T & 1 & 7 &  +12\%
        \end{tabular}
    \label{table:ablation}
    \end{center}
\vspace{-6mm} 
\end{table}

\subsubsection{\textbf{Attention Heads}}

Multihead attention uses k differently-parameterized attention mechanisms to stabilize and improve learning. Best results are found for 3 attention heads. An evident detriment in performance can be seen when using just one head. It is worth noting, that exploration of gradients histograms clearly reveals the prevalence of one  head above the others for the final prediction.

\subsubsection{\textbf{Final Feed-Forward Layer} (FC)}

We add a final fully-connected layer that operates on each node’s extracted features. Table \ref{table:ablation} shows the improvement of using this final layer in prediction performance.



\subsubsection{\textbf{Beta ($\beta$)}} 

We try different weights for the FDE term in the loss function. In this case, we deploy the relative improvement in both ADE and FDE. It is crystal clear that there is a   trade-off between improving FDE and ADE. Higher values of $\beta$ achieve better FDE results, but once  it becomes higher than 1, ADE values are affected.

 \subsubsection{\textbf{Social consistency}}
 
 Finally, we explore how the socially-consistent term added to the  loss function affect the final performance. This term  penalizes the overlap among the estimated output for all the road agents in a sequence. This overlap is computed by means of Bolzano's Theorem, from which we obtain a percentage of overlap. This value is added to the loss function weighted with hyperparameter $\alpha$.  After several tests, we found that the optimum value for $\alpha$ is 5, achieving not only a significant improvement compared to the model trained without this penalty, but also allowing us to attain a lower collision rate.

We also explore different sets of inputs, being our final choice position, heading and type of object, since it clearly showed best performance. Predicting velocity instead of position is found to be beneficial when the prediction horizon is 3s, probably due to the smoother trajectory predictions. Hence, for ApolloScape Challenge, we predict velocities instead of positions. However, for inD benchmark we found better results when predicting directly x,y-coordinates.
In Table \ref{table:agents}, ADE and FDE absolute results for each intersection and agent type are detailed.  \textit{Vehicle} encompasses cars, vans and trucks, \textit{Pedestrian} includes scooters and \textit{Bycicle} contains motorcycles.

\begin{table}[htp]
    \caption{ADE/FDE test results on inD Dataset.}
    \begin{center}
        \begin{tabular}{c||c|c|c||c}
        \textbf{Scenario} & \textbf{Vehicle} & \textbf{Pedestrian} & \textbf{Bycicle} & \textbf{Avg.}  \\ \hline\hline
        Intersection A & 0.67/1.54 & 0.50/1.12 & 0.29/0.65  &  0.67/1.55 \\
        Intersection B &  0.41/0.91 & 0.54/1.21  & 0.46/1.01 &  0.48/1.08 \\
        Intersection C &  0.18/0.41  & 0.52/1.15  & 0.51/1.14 & 0.30/0.69  \\
        Intersection D & 0.39/0.80  &  0.41/0.84 & 0.23/0.48 & 0.40/0.83 \\ \hline
        Average & 0.36/0.80 & 0.52/1.16 & 0.47/1.06 & 0.43/0.98        \end{tabular}
    \label{table:agents}
    \end{center}
    \vspace{-6mm}
\end{table}

\subsection{Experimental Results on InD and Apollo Benchmarks}

\begin{table*}[htp]
    \caption{Comparative Results on InD benchmark measured by ADE/FDE.}
    \begin{center}
        \begin{tabular}{c|cccc|c} 
        \textbf{Method} & Intersection A & Intersection B & Intersection C & Intersection D & Avg.  \\ \hline\hline
        S-LSTM &  2.29 / 5.33 & 1.28 / 3.19 & 1.78 / 4.24 & 2.17 / 5.11 & 1.88 / 4.47 \\
        S-GAN & 3.02 / 5.30 & 1.55 / 3.23 & 2.22 / 4.45 & 2.71 / 5.64 & 2.38 / 4.66\\
        GRIP++ & 1.65 / 3.65 & 0.94 / 2.06 &  0.59 / 1.41 & 1.94 / 4.46 & 1.28 / 2.88 \\ 
        AMENet & 1.07 / 2.22 &  0.65 / 1.46 & 0.83 / 1.87 & 0.37 / 0.80 & 0.73 / 1.59  \\
        DCENet & 0.96 / 2.12 & 0.64 / 1.41 & 0.86 / 1.93 & \textbf{0.28 / 0.62} & 0.69 / 1.52 \\\hline
        \textbf{SCOUT (Ours)} & \textbf{0.67 / 1.55} &  \textbf{0.48 / 1.08} &   \textbf{0.30 / 0.69} & 0.40 / 0.83 &\textbf{ 0.46 / 1.03}
       \end{tabular}
      \end{center}
      \vspace{-2mm}
    \label{table:ind}
    
\end{table*}

\begin{table*}[htp]
    \caption{Comparative Results on ApolloScape Dataset.}
    \centering
        \begin{tabular}{c|c|ccc|c|ccc} 
        
        \textbf{Method} & WSADE & ADEv & ADEp & ADEb & WSFDE & FDEv & FDEp & FDEb  \\ \hline\hline 
        TrafficPredict &  8.58 & 7.94 & 7.18 & 12.88 & 24.23 & 12.77 & 11.12 & 22.79   \\
        S-GAN & 1.96 & 3.15 & 1.33 & 1.86 & 3.59 & 5.66 & 2.45 & 4.72\\
        S-LSTM & 1.89  &  2.95 & 1.29 & 2.53 & 3.40 & 5.28 & 2.32 & 4.54  \\
        StarNet & 1.34 & 2.39 & 0.78 & 1.86 & 2.49 & 4.28 & 1.51 & 3.46  \\
        GRIP++ & 1.27 & 2.24 & \textbf{0.71} & 1.85 & 2.39 & 4.07 & \textbf{1.38} & 3.53\\ \hline
        \textbf{SCOUT (Ours)} &  \textbf{1.26}  & \textbf{2.21 } & 0.73 & \textbf{1.82}  & \textbf{2.35} & \textbf{3.93}  & 1.41  & \textbf{3.37} 
       \end{tabular}
    \label{table:apollo}
  
\vspace{-0.2cm}  
\end{table*}

SCOUT is compared to other baselines and existing solutions using InD dataset following the same data preprocessing strategy as \cite{cheng2020exploring} in order to make a fair comparison. $T_{obs} = 3.2s$ and $T_{pred} = 4.8s$ are used, being the time interval $0.4s$, i.e. 8 frames for observation and 12 for prediction. For each intersection, 1/3 of recordings are kept for testing. 80\% of the remaining recordings are used for training and 20\% for validation. 
We also compare our system against other methods on ApolloScape leaderboard that have publications. 
All used baselines are detailed hereunder: 

\begin{itemize}
    \item \textit{TrafficPredict} \cite{Ma_Zhu_Zhang_Yang_Wang_Manocha_2019}: A long short-term memory-based (LSTM-based) prediction algorithm that uses an instance layer and a category layer to learn trajectories and interactions. It is the baseline of the ApolloScape Trajectory Dataset.
    \item {\textit{Social LSTM (S-LSTM)} \cite{Alahi2016SocialLH}}: uses LSTMs to extract features of trajectory and propose social pooling to model interactions for pedestrian trajectory prediction.
    \item{\textit{Social GAN (S-GAN) }\cite{Gupta2018SocialGS}}: proposes a conditional GAN-based trajectory predictor.
    \item{\textit{StarNet} \cite{Zhu2019StarNetPT}}: ranked  \#1  in  the  CVPR2019  trajectory  prediction  challenge. It uses a star topology which includes a hub networrk that takes observed trajectories to model interactions and multiple host networks, each of which corresponds to one pedestrian for predicting future trajectories.
    \item \textit{GRIP++ }\cite{li2020gripplus}:   the enhanced version of GRIP \cite{DBLP:journals/corr/abs-1907-07792} uses a graph to represent the interactions of close objects, applies several graph convolutional blocks to extract features, and subsequently uses an encoder-decoder LSTM model to make predictions.
    \item{\textit{Attentive Maps Encoder Network (AMENet)} \cite{CHENG2021253}}: a  generative model based on a conditional variational auto-encoder (CVAE) that uses attentive dynamic maps for interaction modeling.
    \item \textit{Dynamic Context Encoder Network (DCENet)} \cite{cheng2020exploring}: the most recent state-of-the-art on the inD benchmark. It extracts spatial context by means of self-attention architectures that is fed along with observed trajectories to two LSTM encoders. Future trajectories are sampled from the latent space encoded by a CVAE.

\end{itemize}

\begin{figure*}
     \centering
      \includegraphics[width=\textwidth]{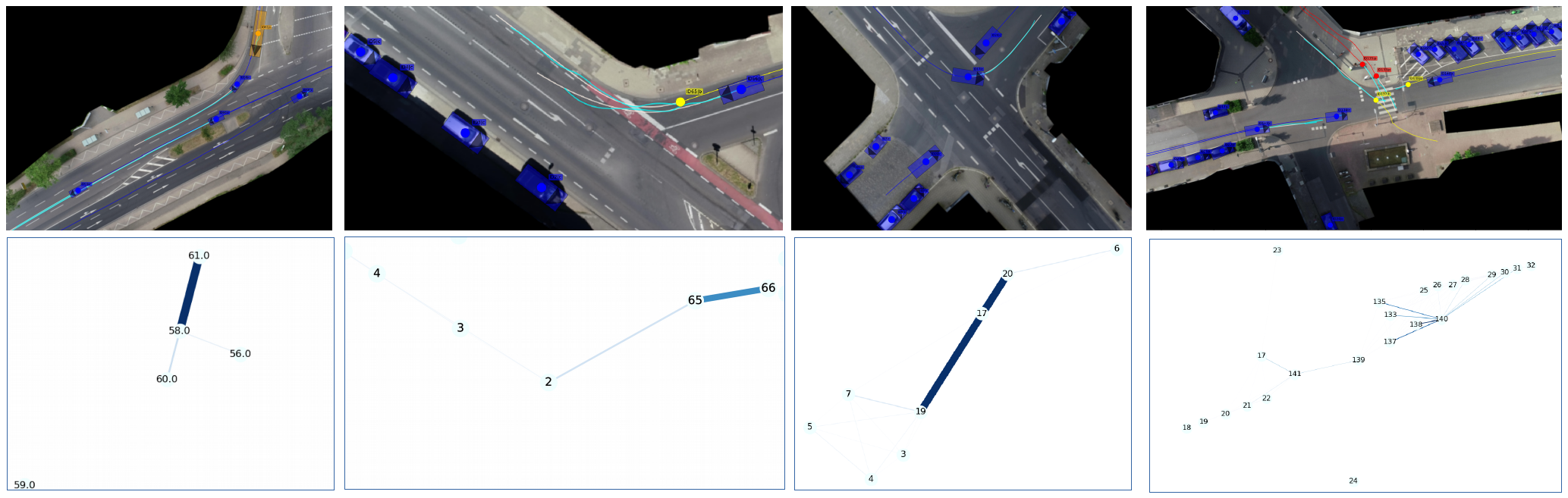}
      \caption{Qualitative results and interaction importance visualization on inD dataset. From left to right: case 1 to 4. }
    \label{fig:interactions}
\end{figure*}

Table \ref{table:ind} summarizes the quantitative results measured by ADE/FDE on inD, where all models were trained and tested using the same data preprocessing. It is clear that SCOUT achieves superior performance by a margin both in ADE and FDE terms. The aforementioned training and testing data splits were chosen to keep it consistent with the results already published on the inD benchmark. Notwithstanding, we believe one of the intersections should be kept for testing, completely unseen during training, in order to make a fair evaluation of the system. In this way, we can assure the model has not overfitted the data by learning trajectories distributions of each of the 4 scenarios. The intersections differ greatly from one to another, since they are located in completely different areas 
with a wide variety of trajectory and interaction distributions, number and type of agents. Hence, this split would be the appropriate to test the generalizability of the model. We perform experiments keeping Intersection D for testing, which amounts to a total of 3 recordings, and Intersection A for validation (7 recordings), leaving the remaining 22 recordings for training. After training the system using this partitioning, we obtain an ADE of 0.56m and FDE of 1.07m, showing that our model is able to predict trajectories in completely new road topologies and scenarios, yet surpassing all baselines. 
Fig. \ref{fig:interactions} depicts some qualitative results across all intersections.

In Table \ref{table:apollo},  results obtained directly from the ApolloScape leaderboard are deployed. SCOUT achieves the best performance in global WSADE and WSFDE. It only slightly fell behind the GRIP++ model on categories ADEp and FDEp. 
These results indicate that our model is able to achieve superior performance on datasets of a completely different nature.

\subsection{Analysis of Scene Transferability}
\label{sec:transfer}

A major challenge in autonomous driving is to find an algorithm that is transferable between different scenarios, i.e. having the capability to generalize beyond the training distribution, which would allow the autonomous vehicle to adapt quickly to previously unseen distributions.  One main advantage of using a generic representation of the scene as a graph is that it allows the model to be transferable between different environments and road topologies. 
In this section, we make use of RounD dataset to explicitly analyse the generalizability of our model, executing two different studies:

\begin{itemize}
    \item First, we trained our model under InD dataset to predict 8 frames from 8 observed frames and test it against an 8-way roundabout from the RounD dataset. First column of Table \ref{table:transferability} shows ADE and FDE values obtained from (I) a regular model trained and evaluated with data from the InD dataset and (II) the zero-shot transferred model learned using training data from the roundabout scenario without additional training on the intersection data. 
    
    \item Secondly, we trained our model using sequences from the RounD dataset   and test it against an unsignalized intersection from the inD dataset. Second column of Table  \ref{table:transferability} deploys the results obtained from (I) a conventional model trained and evaluated using only RounD data and (II) a zero-shot model learned under InD  dataset  without additional training on the roundabout scenario.  
\end{itemize}
\begin{table}[htp]
    \caption{Transferability assesment (ADE/FDE).}
    \begin{center}
        \begin{tabular}{c|c|c}
         & \textbf{InD Dataset} & \textbf{RounD Dataset}  \\ \hline 
        Conventional Model  &  0.33 / 0.52  &  1.22 / 2.55  \\  \hline
        Zero-shot Model & 0.54 / 1.02   &   1.38 / 3.45
        \end{tabular}
    \label{table:transferability}
    \end{center}
    \vspace{-5mm}
\end{table}
There are some conclusions that can be drawn from these results. On the one hand, performance in RounD is noticeably worse due to two main factors.  First, it includes trajectories that are significantly more challenging than those from inD dataset given the topology of the roundabouts  and the plentiful interactions that happen among agents that move at high speeds. Secondly, the amount of sequences is half the number of sequences obtained from inD dataset. On the other hand,  the model trained under RounD shows better results when tested against an inD scenario. This proves our hypothesis that RounD trajectories are more difficult to predict and that our model is able to generalize. When testing a transferred model from inD to RounD, there is a remarkable change in performance depending on the specific scenario used for testing. The first two scenarios correspond to smaller roundabouts where there exist more linear trajectories. In this case, the model trained under inD performs better than the conventional model. However, when we use larger and more challenging roundabouts for testing, the opposite results are found. Table \ref{table:transferability} shows the average values of ADE and FDE.

\subsection{Visualization of the influence of interactions.}

One serious concern when dealing with neural networks is their lack of interpretability or explainability due to the black-box nature of most high-performance state-of-the-art models. 
Therefore, one of our goals in this work was to keep the model simple enough to be able to carefully explore it. In this section, we provide some insights that help us better explain our model's predictions. 
Given that the gradient is the derivative of the loss function with respect to a particular parameter, parameters for which a small change could drastically change the loss of an example are expected to have a large derivative relative to that of a non-relevant parameter. Following this intuition, we chose a particular technique called\textit{ Integrated Gradients} \cite{sundararajan2017axiomatic} that attributes a given model's prediction to input features, relative to a certain baseline input. 
More concretely, Integrated Gradients  represents the integral of gradients with respect to inputs along the path from a given baseline to input. 
In our case, we use edge weights   as inputs to understand the influence of each interaction in the final prediction.  To illustrate the results found, we plot the matching graph that is fed to the model for each of the scenes in Fig. \ref{fig:interactions}. In this visualization, edge colors and thickness represent the importance of each interaction for the final prediction. Following this procedure, we analyse four scenarios:

\begin{itemize}
    \item \textit{Case 1}: T-junction in an industrial area with few interactions. Here, the only interaction that has some impact on the final prediction is the one between the truck and the car situated in front of it.
    \item \textit{Case 2}: a T-junction, where the main road contains cycle paths. In this scenario we can observe a cyclist and a car behind of it aiming to turn right, being the connection between both agents the strongest one.
    \item \textit{Case 3}: A four-armed intersection with a priority road where one vehicle is stopped at the give way sign yielding to oncoming traffic. The corresponding graph clearly shows the importance of the interaction among the three vehicles involved.
    \item \textit{Case 4}: city center scenario with a zebra crossing near a 4-way intersection, where two cyclists, two pedestrians and two vehicles approach the zebra crossing. 
\end{itemize}

We can also directly visualize the attention weight between each pair of nodes in order to understand the attention learned. Because this weight is associated with edges, we can visualize it in the same way as with Integrated Gradients. As expected, we obtain similar results from both techniques. We also found connections with parked cars, as shown in \textit{Case 2} and \textit{Case 4}. We believe these cars are seen as road limits, and hence are also important for predicting trajectories of moving agents.


\section{Conclusions}
 \label{sec:conclusion}
 
 In this work, we propose SCOUT, a novel attention-based graph neural network for socially-aware and socially-consistent trajectory forecasting. Our scheme obtains a flexible and generic high-level representation of the scene as a graph by modelling each agent as a node and interactions as edges. Agent's features are aggregated using an attention mechanism and the extracted features are fed to a final feed-forward network for the final behavior prediction. The simplistic nature of our model allows us to explore it in order to understand the importance of each interaction for the final prediction. We also perform an extensive ablation study to understand the contribution of each component to the overall system and analyse its flexibility and transferability by testing it against a completely new scenario. Our scheme achieves state-of-the-art results in both ApolloScape and inD benchmarks. Future work will explore the prediction of intentions rather than trajectories and the contribution of using image data as additional input to our system.
 

\section{Acknowledgements}

 This work has been funded by research grants S2018/EMT-4362 SEGVAUTO 4.0-CM (Community  Region of Madrid) and DPI2017-90035-R  (Spanish Ministry of Science and Innovation).

\bibliographystyle{IEEEtran}
\bibliography{references}
\end{document}